\documentclass{llncs}
\usepackage{algorithm}
\usepackage{algpseudocode}
\usepackage{graphicx}
\usepackage{url}
\usepackage{xspace}
\usepackage{hyperref}
\usepackage{color}
\usepackage{array}
\usepackage{underscore}
\usepackage{tipa}
\usepackage{cite}
\usepackage{stmaryrd}
\usepackage{amsmath}
\usepackage{amssymb}
\usepackage{booktabs}
\usepackage{listings}
\usepackage{paralist}
\usepackage{subfig}
\usepackage{hyperref}
\usepackage{breakurl}
\usepackage{tikz}
\usepackage{textcomp}
\usepackage{alltt}
\usepackage{amsmath}

\algrenewcommand\algorithmicrequire{\textbf{Input}}
\algrenewcommand\algorithmicensure{\textbf{Output}}

\usepackage{footnote}

\makeatletter

\def\holdocspecials{\do\ \do\$\do\&%
  \do\#\do\^\do\^^K\do\_\do\^^A\do\%}

\def\holtt{\trivlist \item[]\if@minipage\else\vskip\parskip\fi
\leftskip\@totalleftmargin\rightskip\z@
\parindent\z@\parfillskip\@flushglue\parskip\z@
\@tempswafalse \def\par{\if@tempswa\hbox{}\fi\@tempswatrue\@@par}
\obeylines \tt \let\do\@makeother \holdocspecials
 \frenchspacing\@vobeyspaces}

\makeatother

\newlength{\hsbw}
\newcommand\HOLSpacing{13pt}

   \newcommand\hilbert{\varepsilon}

   \newcommand{\Cond}{\(\rightarrow\)}
   \newcommand{\Eqv}{\(\equiv\)}
   \newcommand{\Iff}{\(\Longleftrightarrow\)\hspace{-1.5mm}}
   \newcommand{\Fa}{\(\forall\)}
   \newcommand{\Et}{\(\exists\)}
   \newcommand{\Eu}{\(\exists_{unique}\)}

   \newcommand{\Impl}{\(\Longrightarrow\)\hspace{-1.5mm}}
   \newcommand{\Func}{\(\to\)\hspace{-1.5mm}}

   \newcommand{\Lam}{\(\lambda\)}
   
   \newcommand{\Minus}{\(-\)}
   \newcommand{\Lminus}{\(-\)\hspace{-1.5mm}}
   \newcommand{\Prime}{\('\)}
   \newcommand{\Und}{\_}
   \newcommand{\Lt}{\(<\)}
   \newcommand{\Gt}{\(>\)}
   \newcommand{\Leq}{\(\leq\)}
   \newcommand{\Geq}{\(\geq\)}
   \newcommand{\Eq}{\(=\)}
   \newcommand{\Lrb}{\((\)}
   \newcommand{\Rrb}{\()\)}

   \newcommand{\Next}{\(\bigcirc\)}
   \newcommand{\Prev}{\(\ominus\)}
   \newcommand{\WPrev}{\(\widetilde{\bigcirc}\)}
   \newcommand{\Event}{\(\Diamond\)}
   \newcommand{\Once}{\(\underline{\Diamond}\)}  
\newcommand{\Hilbert}{\(\hilbert\)}

\newcommand{\Conj}{\(\wedge\)}
\newcommand{\Disj}{\(\vee\)}
\newcommand{\Neg}{\(\neg\)}
\newcommand{\Pnd}{\(\Diamond\)}

\newcommand{\Models}{\(\models\)}

\long\def\rechol#1#2#3{\let\next=\rechol\def\postnext{#2#3}\ifx#1\end
\let\next=\relax\def\postnext{\relax}
\else\ifx#1!\Fa                                          
\else\ifx#1@\Hilbert                                     
\else\ifx#1\#\Pnd                                        
\else\ifx#1'\Prime                                       
\else\ifx#1~\Neg                                         
\else\ifx#1\~\Neg
\else\ifx#1_\Und                                         
\else\ifx#1(\ifx#2+\ifx#3)\Next\def\postnext{}\fi        
            \else\ifx#2-\Prev\def\postnext{}             
            \else\ifx#2~\ifx#3)\WPrev\def\postnext{}\fi            
             \else\Lrb\fi\fi\fi                          
\else\ifx#1)\Rrb%
\else\ifx#1\/\Disj                                       
\else\ifx#1\.\Lam                                        
\else\ifx#1>\ifx#2=\Geq\def\postnext{#3}\else\Gt\fi      
\else\ifx#1?\ifx#2!\Eu\def\postnext{#3}\else\Et\fi       
\else\ifx#1-\ifx#2>\Func\def\postnext{#3}               
            \else\ifx#2-\Lminus\def\postnext{#3}
            \else\Minus\fi\fi                               
\else\ifx#1|\ifx#2-\Turns\def\postnext{#3}               
            \else\ifx#2=\Models\def\postnext{#3}
                 \else\Bar\fi\fi
\else\ifx#1<\ifx#2=\ifx#3>\Iff\def\postnext{}       
                   \else\Leq\def\postnext{#3}\fi    
            \else\ifx#2+\Event\def\postnext{}       
            \else\ifx#2-\Once\def\postnext{}       
            \else\Lt\fi\fi\fi                       
\else\ifx#1=\ifx#2=\ifx#3>\Impl\def\postnext{}            
                   \else\Eqv\def\postnext{#3}\fi         
            \else\ifx#2>\Cond\def\postnext{#3}
                 \else\Eq\fi\fi
\else\ifx#1/\ifx#2\^^M\Conj\par\def\postnext{#3}         
            \else\ifx#2\ \Conj\ \def\postnext{#3}\else#1\fi\fi  
\else#1\fi\fi\fi\fi\fi\fi\fi\fi\fi\fi\fi\fi\fi\fi\fi\fi\fi\fi\fi
\expandafter\next\postnext}

\def\systemname#1{\textsf{#1}\xspace}
\newcommand{\coloneqq}[0]{\mathrel{\mathop:}=}
\newcommand{\HOLLight}{\systemname{HOL Light}}

\title{Lemma Mining over \HOLLight }
\author{Cezary Kaliszyk \and Josef Urban}

\institute{ University of Innsbruck, Austria \and Radboud University, Nijmegen}

\makeatletter
\renewcommand\section{\@startsection{section}{1}{\z@}%
                       {-12\p@ \@plus -4\p@ \@minus -4\p@}%
                       {8\p@ \@plus 4\p@ \@minus 4\p@}%
                       {\normalfont\large\bfseries\boldmath
                        \rightskip=\z@ \@plus 8em\pretolerance=10000 }}
\makeatother

\begin{document}
\maketitle

\begin{abstract}
Large formal mathematical libraries consist of millions of atomic
inference steps that give rise to a corresponding number of proved
statements (lemmas). Analogously to the informal mathematical
practice, only a tiny fraction of such statements is named and
re-used in later proofs by formal mathematicians. In this work,
we suggest and implement criteria defining the estimated usefulness
of the HOL Light lemmas for proving further theorems. We use these
criteria to mine the large inference graph of all lemmas in the core
HOL Light library, adding thousands of the best lemmas to the pool
of named statements that can be re-used in later proofs. The
usefulness of the new lemmas is then evaluated by comparing the
performance of automated proving of the core HOL Light theorems with
and without such added lemmas.

\end{abstract}

\section{Introduction}
\label{Introduction}

In the last decade, large formal mathematical corpora such as the
Mizar Mathematical Library~\cite{mizar-in-a-nutshell} (MML), Isabelle/HOL~\cite{WenzelPN08} and HOL Light~\cite{Harrison96}/Flyspeck~\cite{Hales05}
have been translated to formats that allow easy experiments with
external automated theorem provers (ATPs) 
and AI systems~\cite{Urb04-MPTP0,MengP08,holyhammer}. 
Several AI/ATP methods for reasoning in the context of a
large number of related theorems and proofs have been suggested and
tried already, including: (i) methods (often external to the core
ATP algorithms) that select relevant premises (facts)
from the thousands of theorems available in such corpora~\cite{HoderV11,KuhlweinLTUH12}, (ii) methods
for internal guidance of ATP systems when reasoning in the large-theory
setting~\cite{UrbanVS11}, (iii) methods that automatically evolve more and more
efficient ATP strategies for the clusters of related problems from
such corpora~\cite{blistr}, and (iv) methods that learn which of such specialized strategies to use
for a new problem~\cite{KuhlweinSU13}.

In this work, we start to complement the first set of methods --
ATP-external premise selection -- with \textit{lemma mining} from the large corpora. The main
idea of this approach is to enrich the pool of human-defined main
(top-level) theorems in the large libraries with the most
useful/interesting lemmas extracted from the proofs in these
libraries. Such lemmas are then eligible together with (or instead of)
the main library theorems as the premises that are given to the ATPs
to attack new conjectures formulated over the large libraries. 

This high-level
idea is straightforward, but there are a number of possible
approaches involving a number of issues to be solved, starting 
with a reasonable definition of a \textit{useful/interesting lemma},
and with making such definitions efficient over corpora that contain
millions to billions of candidate lemmas. These issues are discussed
in Sections~\ref{Data} and \ref{Good}, after motivating and explaining the overall approach 
for using lemmas in large theories in Section~\ref{Approach} and giving an overview of the 
recent related work in Section~\ref{Related}.

As in any AI discipline dealing with large amount of data, research in
the large-theory field is driven by rigorous experimental evaluations
of the proposed methods over the existing corpora. 
For the first experiments with lemma
mining we use the HOL Light system, together with its core library and the Flyspeck library.
The various evaluation scenarios are defined and discussed in Section~\ref{Scenarios},
and the implemented methods are evaluated
in Section~\ref{Experiments}.
Section~\ref{Future} discusses the
various future directions and concludes.

\section{Using Lemmas for Theorem Proving in  Large Theories}
\label{Approach}
The main task in the Automated Reasoning in Large Theories (ARLT) domain is to prove new
conjectures with the knowledge of a large body of previously proved
theorems and their proofs. This setting reasonably corresponds to how
large ITP libraries are constructed, and hopefully also emulates how human mathematicians work more
faithfully than the classical scenario
of a single hard problem consisting of isolated axioms and a
conjecture~\cite{UrbanV13}. The pool of previously proved theorems
ranges from thousands in large-theory ATP benchmarks such as MPTP2078~\cite{abs-1108-3446},
to tens of thousands when working with the whole ITP
libraries.\footnote{14185 theorems are in the HOL/Flyspeck library, about 20000 are in the Isabelle/HOL library, and
  about 50000 theorems are in the Mizar library.}

The strongest existing ARLT systems combine variously parametrized
premise-selection techniques (often based on machine learning from previous proofs) with  
ATP systems and their strategies that are called with varied
numbers of the most promising premises. These techniques can go quite
far already: when using 14-fold parallelization and 30s wall-clock
time, the HOL(y)Hammer system~\cite{holyhammer,KaliszykU13} can today prove
47\% of the 14185 Flyspeck theorems~\cite{EasyChair:74}. This is
measured in a scenario\footnote{A similar scenario has been 
  introduced in 2013 also for the CASC LTB competition.} in which the Flyspeck
theorems are ordered \textit{chronologically} using the loading
sequence of the Flyspeck library, and presented in this order to
HOL(y)Hammer as conjectures. After each theorem is attempted, its human-designed HOL
Light proof is fed to the HOL(y)Hammer's learning components, together
with the (possibly several) ATP proofs found by HOL(y)Hammer
itself. This means that for each Flyspeck theorem, all human-written
HOL Light proofs of all previous theorems are assumed to be known,
together with all their ATP proofs found already by HOL(y)Hammer, but
nothing is known about the current conjecture and the following parts
of the library (they do not exist yet).

So far, systems like HOL(y)Hammer (similar systems include Sledgehammer/MaSh~\cite{KuhlweinBKU13} 
and MaLARea~\cite{US+08}) have only used the set of \textit{named library theorems} for
proving new conjectures and thus also for the premise-selection
learning. This is usually a reasonable set of theorems to start with,
because the human mathematicians have years of experience with
structuring the formal libraries. On the other hand, there is no
guarantee that this set is in any sense optimal, both for the human
mathematicians and for the ATPs. The following three
observations indicate that the set of human-named theorems may be suboptimal:
\begin{itemize}
\item[\textit{Proofs of different length:}] The human-named theorems may differ
  considerably in the length of their proofs. The human naming is based
  on a number of (possibly traditional/esthetical) criteria that may
  sometimes have little to do with a good structuring of the library.
\item[\textit{Duplicate and weak theorems:}] The large collaboratively-build
  libraries are hard to manually guard against duplications and naming
  of weak versions of various statements. The experiments with the
  MoMM system over the Mizar library~\cite{Urban06-ijait} and with the recording of the Flyspeck library~\cite{KaliszykK13} have shown
  that there are a number of subsumed and duplicated theorems, and that some
  unnamed strong lemmas are proved over and over again.
\item[\textit{Short alternative proofs:}] The experiments with
  AI-assisted ATP over the Mizar and Flyspeck
  libraries~\cite{AlamaKU12,holyhammer} have shown that the combined
  AI/ATP systems may sometimes find alternative proofs that are much
  shorter and very different from the human proofs, again turning some
  ``hard'' named theorems into easy corollaries.
\end{itemize}

Suboptimal naming may obviously influence the performance of the
current large-theory systems.
If many important lemmas are
omitted by the human naming, the ATPs will have to find them over and
over when proving the conjectures that depend on such lemmas. On the
other hand, if many similar variants of one theorem are named, the
current premise-selection methods might focus too much on those
variants, and fail to select the complementary theorems that are also
necessary for proving a particular conjecture.\footnote{This behavior
  obviously depends on the premise-selection algorithm. It
  is likely to occur when the premise selection is mainly based on
  symbolic similarity of the premises to the conjecture. It is less
  likely to occur when complementary semantic selection criteria are additionally
  used as, e.g.,  in SRASS~\cite{SutcliffeP07} and MaLARea~\cite{US+08}.}

To various extent, this problem might be remedied by the
alternative learning/guidance methods (ii) and (iii) mentioned in the
introduction: Learning of internal ATP guidance using for example
Veroff's hint technique~\cite{Veroff96}, and learning of suitable ATP strategies
using systems like BliStr~\cite{blistr}. But these methods are so far much more
experimental in the large-theory setting than premise
selection.\footnote{In particular, several initial experiments done so
  far with Veroff's hints %
over the MPTPChallenge and
  MPTP2078 benchmarks were so far unsuccessful.} That is why we
propose the following lemma-mining approach:

\begin{enumerate}
\item Considering (efficiently) the detailed graph of all atomic inferences
  contained in the ITP libraries. Such a graph has millions of nodes
  for the core HOL Light corpus, and hundreds of millions of nodes for the whole Flyspeck.
\item Defining over such large proof graphs efficient criteria 
  that select a smaller set of the strongest
  and most orthogonal lemmas from the corpora.
\item Using such lemmas together with (or instead of) the human-named
  theorems for %
proving new conjectures
  over the %
corpora.
\end{enumerate}

\section{Overview of Related Work and Ideas }
\label{Related}

A number of ways how to measure the quality of lemmas and how to use
them for further reasoning have been proposed already, particularly in
the context of ATP systems and proofs.  
Below we summarize recent approaches and tools that initially seemed most relevant to our work.

Lemmas are an essential part of various ATP
algorithms. State-of-the-art ATPs such as Vampire~\cite{Vampire}, E~\cite{Sch02-AICOMM} and Prover9~\cite{McC-Prover9-URL}
implement various variants of the ANL loop~\cite{WO+84}, resulting in hundreds to
billions of lemmas inferred during the prover runs. This gave rise to
a number of efficient ATP indexing techniques, redundancy control
techniques such as subsumption, and also fast ATP heuristics (based on
weight, age, conjecture-similarity, etc.) for choosing the best lemmas
for the next inferences. Several ATP methods and tools work with
such ATP lemmas.  Veroff's \textit{hint technique}~\cite{Veroff96} extracts the best
lemmas from the proofs produced by successful Prover9 runs and uses
them for directing the proof search in Prover9 on related problems. A
similar lemma-extracting, generalizing and proof-guiding technique (called \textit{E Knowledge Base -- EKB}) was
implemented by Schulz in E prover as a part of his PhD thesis~\cite{Sch00}. 

Schulz also implemented the \textit{epcllemma} tool that estimates the
best lemmas in an arbitrary DAG (directed acyclic graph) of inferences. 
Unlike the hint-extracting/guiding 
methods, this tool works not
just on the handful of lemmas involved in the final refutational
proof, but on the typically very large number of lemmas produced
during the (possibly unfinished) ATP runs.  The epcllemma's criteria
for selecting the next best lemma from the inference DAG are: (i) the size of
the lemma's inference subgraph based at the nodes that are either axioms
or already chosen (better) lemmas, and (ii) the weight of the
lemma. This lemma-selection process may be run recursively, until a stopping
criterion (minimal lemma quality, required number of lemmas, etc.) is
reached. Our algorithm for HOL Light (Section~\ref{Good}) is
quite similar to this.

\textit{AGIntRater}~\cite{PuzisGS06} is a tool that computes various characteristics of the
lemmas that are part of the final refutational ATP proof and aggregates them into
an overall \emph{interestingness} rating. These characteristics include:
obviousness, complexity, intensity, surprisingness, adaptivity, focus,
weight, and usefulness, see~\cite{PuzisGS06} for details. AGIntRater so
far was not directly usable on our data for various reasons
(particularly the size of our graph), but we might re-use and try to
efficiently implement some of its ideas later.

Pudl\'{a}k~\cite{Pud06-ESCoR} has conducted experiments over several datasets
with automated re-use of lemmas from many existing ATP proofs in order to find
smaller proofs and also to attack unsolved problems. This is similar
to the hints technique, however more automated and closer to our large-theory setting
(hints have so far been successfully applied mainly in small algebraic
domains). To interreduce the large number of such lemmas with respect
to subsumption he used the E-based \textit{CSSCPA}~\cite{Sut01-LPAR} subsumption tool by Schulz
and Sutcliffe. \textit{MoMM}~\cite{Urban06-ijait} adds a number of large-theory features to
CSSCPA. It was used for (i) fast interreduction of million of lemmas extracted (generalized)
from the proofs in the Mizar library, and (ii) as an early ATP-for-ITP hammer-style tool for
completing proofs in Mizar with the help of the whole Mizar library. All
library lemmas can be loaded, indexed and considered for each query, 
however the price for this breadth of coverage is
that the inference process is limited to subsumption extended with
Mizar-style dependent types. 

AGIntRater and epcllemma use a lemma's position in the inference graph
as one of the lemma's characteristics that contribute to its
importance. There are also purely graph-based algorithms that try to
estimate a relative importance of nodes in a graph. In particular,
research of large graphs became popular with the appearance of the
World Wide Web and social networks. Algorithms such as
\textit{PageRank}~\cite{page98pagerankTechreport} (eigenvector
centrality) have today fast approximative implementations that easily
scale to billions of nodes.

\section{The Proof Data}
\label{Data}

We initially consider two corpora: the core HOL Light corpus (SVN version 146) and the
Flyspeck corpus (SVN version 2886). The core HOL Light corpus consists of 1,984 named
theorems, while the Flyspeck corpus contains 14,185 named
theorems. There are 97,714,465 lemmas in Flyspeck when exact duplicates are
removed, and 420,253,109 lemmas when counting duplicates. When removing duplicates only within the proof of each named theorem, the final number of lemmas is 146,120,269.
 For
core HOL Light the number of non-duplicate lemmas is 1,987,781. When counting duplicates it is 6,963,294, and when removing duplicates only inside the proof of each named theorem it is 2,697,212 .  To
obtain the full inference graph for Flyspeck we run the
proof-recording version of HOL Light~\cite{KaliszykK13}. This takes 14
hours of CPU time and 42 GB of RAM on an Intel Xeon 2.6 GHz
machine. This time and memory consumption are much lower when working
only with the core HOL Light, hence many of the experiments were so far
done only on the smaller corpus.

There are 140,534,426 inference edges between the unique Flyspeck lemmas,
each of them corresponding to one of the LCF-style  kernel inferences done by HOL
Light~\cite{KaliszykK13}.  During the proof recording we additionally
export the information about the symbol weight (size) of each lemma,
and its normalized form that serially numbers bound and free variables
and tags them with their types. This information is later used for
external postprocessing, together with the information about which theorems where originally named. 
Below is a commented example of the initial segment of
the Flyspeck proof trace, the full trace (1.5G in size) is available
online\footnote{\url{http://mizar.cs.ualberta.ca/~mptp/lemma_mining/proof.trace.old.gz}}, as well as the numbers of the original named Flyspeck theorems.\footnote{\url{http://mizar.cs.ualberta.ca/~mptp/lemma_mining/facts.trace.old.gz}}

\begin{small}
\begin{verbatim}
F13      #1, Definition (size 13): T <=> (\A0. A0) = (\A0. A0)
R9       #2, Reflexivity (size 9): (\A0. A0) = (\A0. A0)
R5       #3, Reflexivity (size 5): T <=> T
R5       #4, Reflexivity (size 5): (<=>) = (<=>)
C17 4 1  #5, Application(4,1):     (<=>) T = (<=>) ((\A0. A0) = (\A0. A0))	 
C21 5 3  #6, Application(5,3):     (T <=> T) <=> (\A0. A0) = (\A0. A0) <=> T
E13 6 3  #7, EQ_MP(6,3) (size 13): (\A0. A0) = (\A0. A0) <=> T
\end{verbatim}					 
\end{small}                                                           

\subsection{Initial Post-processing and Optimization of the Proof Data}

During the proof recording, only exact duplicates are easy to
detect. Due to various implementational issues, it is simpler to
always limit the duplication detection to the lemmas derived within a
proof of each named theorem, hence this is our default initial
dataset.  HOL Light does not natively use de Bruijn indices for
representing variables, i.e., two alpha-convertible versions of the
same theorems will be kept in the proof trace if they differ in
variable names. Checking for alpha convertibility during the proof
recording is nontrivial, because in the HOL Light's LCF-style approach
alpha conversion itself results in multiple kernel inferences. That is
why we keep the original proof trace untouched, and implement its
further optimizations as external postprocessing of the trace.

In particular, to merge alpha convertible lemmas in a proof trace $T$,
we just use the above mentioned normalized-variable representation of
the lemmas as an input to an external program that produces a new
version of the proof trace $T'$.  This program goes through the trace $T$
and replaces references to each lemma by a reference to the earliest
lemma in $T$ with the same normalized-variable representation. The proofs of
the later named alpha variants of the lemmas in $T$ are however still kept in the new
trace $T'$, because such proofs are important when computing the
usage and dependency statistics over the normalized lemmas.  So far we
have done this postprocessing only for the core HOL Light 2,697,212
lemmas,\footnote{\url{http://mizar.cs.ualberta.ca/~mptp/lemma_mining/human.gz}}
because printing out of the
variable-normalized version of the 146,120,269 partially de-duplicated Flyspeck lemmas would
produce more than 100G of data.  From the 2,697,212 partially de-duplicated core HOL Light lemmas
1,076,995 are left after this stronger normalization.
It is clear that such post-processing operations can %
be
implemented different ways. In this case, some original
information about the proof graph is lost, while some information
(proofs of duplicate lemmas) is still kept, even though it could be also
pruned from the graph, producing a differently normalized
version. 

The ATP experiments described below use only the two versions
of the proof trace described above, but we have also explored some
other normalizations. A particularly interesting optimization from the
ATP point of view is the removal of subsumed lemmas. An initial
measurement with the (slightly modified) MoMM system done on the
clausified first-order versions of about 200,000 core HOL Light lemmas has
shown that about 33\% of the clauses generated from the lemmas are
subsumed. But again, ATP operations like subsumption interact with the
level of inferences recorded by the HOL Light kernel in nontrivial
ways. It is an interesting task to define exactly how the original
proof graph should be transformed with respect to such operations, and
how to perform such proof graph transformations efficiently over the whole
Flyspeck.

\section{Selecting Good Lemmas}
\label{Good}

Several approaches to defining the notion of a useful/interesting
lemma are mentioned in Section~\ref{Related}. There are a number of
ideas that can be explored and combined together in various ways, but
the more complex methods (such as those used by AGIntRater) are not
yet directly usable on the large ITP datasets that we have. So far, we
have experimented mainly with the following techniques:

\begin{enumerate}
\item A direct OCAML implementation of lemma quality metrics based on the HOL
  Light proof-recording data structures. 
\item Schulz's epcllemma and its minor modifications.
\item PageRank, applied in various ways to the proof trace. 
\end{enumerate}

\subsection{Direct Computation of Lemma Quality}
The advantage of the direct OCAML implementation is that no export to external tools is
necessary and all the information collected about the lemmas by the
HOL Light proof recording is directly available.  The basic factors
that we use so far for defining the quality of a lemma $i$ are its:
(i) set of direct proof dependencies $d(i)$ given by the proof trace,
(ii) number of recursive dependencies $D(i)$, (iii) number of
recursive uses $U(i)$, and (iv) number of HOL symbols (HOL weight)
$S(i)$. When recursively defining $U(i)$ and $D(i)$ we assume that in
general some lemmas may already be named ($k \in Named$) and some
lemmas are just axioms ($k \in Axioms$). Note that in HOL Light there
are many lemmas that have no dependencies, but formally they are still
derived using for example the reflexivity inference rule (i.e., we do
not count them among the HOL Light axioms). The recursion when defining $D$ thus
stops at axioms, named lemmas, and lemmas with no dependencies. The
recursion when defining $U$ stops at named lemmas and unused lemmas. Formally:

\begin{definition}[Recursive dependencies and uses]
\begin{align*}
D(i) &=
\begin{cases}
1 & \text{if $i \in Named \lor i \in Axioms$},\\
\sum\limits_{j \in d(i)} D(j) & \text{otherwise.}\\
\end{cases}
\\
U(i) &=
\begin{cases}
1 & \text{if $i \in Named$,}\\
\sum\limits_{i \in d(j)} U(j) & \text{otherwise.}
\end{cases}
\end{align*}
\end{definition}
In particular, this means that 
$$D(i) = 0 \iff d(i) = \emptyset \land \lnot (i \in Axioms)$$ 
and also that
$$U(i) = 0 \iff \forall j \lnot (i \in d(j))$$
These basic characteristics are combined into the following lemma
quality metrics $Q_1(i)$, $Q_2(i)$, and $Q_3(i)$. $Q_1^r(i)$ is a
generalized version of $Q_1(i)$, which we (apart from $Q_1$) test for $r \in \{0, 0.5, 1.5, 2 \}$:

\begin{definition}[Lemma quality]
\begin{equation*}
\begin{aligned}
Q_1(i) &= \frac{U(i) * D(i)}{S(i)}\\
Q_2(i) &= \frac{U(i) * D(i)}{S(i)^2}\\
\end{aligned}
\qquad
\begin{aligned}
Q_1^r(i) &= \frac{U(i)^r * D(i)^{2-r}}{S(i)}\\
Q_3(i) &= \frac{U(i) * D(i)}{1.1^{S(i)}}\\
\end{aligned}
\end{equation*}
\end{definition}
The justification behind these definitions are the following heuristics: 
\begin{enumerate}
\item The higher is $D(i)$, the more necessary it is to remember the lemma $i$, because it will be harder to infer with an ATP when needed.
\item The higher is $U(i)$, the more useful the lemma $i$ is for proving other desired conjectures.
\item The higher is $S(i)$, the more complicated the lemma $i$ is in
  comparison to other lemmas. In particular, doubled size may often
  mean in HOL Light that $i$ is just a conjunction of two other
  lemmas.\footnote{The possibility to create conjunctions is quite a
    significant difference to the clausal setting handled by
    the existing tools. A longer clause is typically weaker, while
    longer conjunctions are stronger. A dependence on a longer
    conjunction should ideally be treated by the evaluating heuristics
    as a dependence on the multiple conjuncts.}
\end{enumerate}

\subsection{Lemma Quality via epcllemma}

Lemma quality in epcllemma is defined on clause inferences
recorded using E's native PCL protocol. The lemma quality computation
also takes into account the lemmas that have been already named, and with minor implementational variations it can be expressed using $D$ and $S$ as follows:
\begin{equation*}
EQ_1(i) = \frac{D(i)}{S(i)}\\
\end{equation*}
The difference to $Q_1(i)$ is that $U(i)$ is not used, i.e., only the
cumulative effort needed to prove the lemma counts, together with its
size (this is also very close to $Q_1^r(i)$ with $r = 0$).  The main advantage of using epcllemma is its fast and robust
implementation using the E code base. This allowed us to load in
reasonable time (about one hour) the whole Flyspeck proof trace into
epcllemma, taking 67 GB of RAM.  Unfortunately, this experiment showed
that epcllemma assumes that $D$ is always an integer. This is likely
not a problem for epcllemma's typical use, but on the Flyspeck graph
this quickly leads to integer overflows and wrong results. To a
smaller extent this shows already on the core HOL Light proof
graph. A simple way how to prevent the overflows was to modify epcllemma to use instead of $D$ the longest chain of inferences $L$:
\begin{equation*}
L(i) =
\begin{cases}
1 & \text{if $i \in Named \lor i \in Axioms$},\\
max_{j \in d(i)} (1+L(j)) & \text{otherwise.}\\
\end{cases}
\end{equation*}
This leads to:
\begin{equation*}
EQ_2(i) = \frac{L(i)}{S(i)}\\
\end{equation*}
Apart from this modification, only minor changes were needed to make
epcllemma work on the HOL Light data. The HOL proof trace was
expressed as a PCL proof (renaming the HOL inferences into E
inferences), and artificial TPTP clauses of the corresponding size
were used instead of the original HOL clauses.

\subsection{Lemma Quality via PageRank}
PageRank (eigenvector centrality of a graph) is a method that assigns
weights to the nodes in an arbitrary directed graph (not just DAG)
based on the weights of the neighboring nodes (``incoming links''). In
more detail, the weights are computed as the dominant eigenvector of
the following set of equations:
\begin{equation*}
PR_1(i) = \frac{1 - f}{N} + f \sum\limits_{i \in d(j)}\frac{PR_1(j)}{|d(j)|}\\
\end{equation*}
where $N$ is the total number of nodes and $f$ is a damping factor,
typically set to 0.85. The advantage of using PageRank is that there
are fast approximative implementations that can process the whole
Flyspeck proof graph in about 10 minutes using about 21 GB RAM, and
the weights of all nodes are computed simultaneously in this time.

This is however also a disadvantage in comparison to the previous
algorithms: PageRank does not take into account the lemmas that have
already been selected (named). The closer a lemma $i$ is to an
important lemma $j$, the more important $i$ will be. Modifications
that use the initial PageRank scores for more advanced clustering
exist~\cite{AvrachenkovDNPS08} and perhaps could be used to mitigate
this problem while still keeping the overall processing reasonably
fast. Another disadvantage of PageRank is its ignorance of the lemma
size, which results in greater weights for the large conjunctions that
are used quite often in HOL Light. $PR_2$ tries to counter that:
\begin{equation*}
PR_2(i) = \frac{PR_1(i)}{S(i)}\\
\end{equation*}
$PR_1$ and $PR_2$ are based on the idea that a lemma is important if
it is needed to prove many other important lemmas. This can be again
turned around: we can define that a lemma is important if it
depends on many important lemmas. This is equivalent to computing the
reverse PageRank and its size-normalized version:
\begin{equation*}
PR_3(i) = \frac{1 - f}{N} + f \sum\limits_{i \in u(j)}\frac{PR_3(j)}{|u(j)|}\\
\qquad
PR_4(i) = \frac{PR_3(i)}{S(i)}\\
\end{equation*}
where $u(j)$ are the direct uses of the lemma $j$, i.e., $i \in u(j) \iff j \in d(i)$. The two ideas can again be combined (note that the sum of the PageRanks of all nodes is always 1):
\begin{equation*}
PR_5(i) = \frac{PR_1(i) + PR_3(i)}{S(i)}\\
\end{equation*}

\subsection{Selecting Many Lemmas}

From the methods described above, only the various variants of
PageRank ($PR_i$) produce the final ranking of all lemmas in one
run. Both epcllemma ($EQ_i$) and our custom methods ($Q_i$) are
parametrized by the set of lemmas ($Named$) that have already been
named. When the task is to choose a predefined number of the best
lemmas, this naturally leads to the following recursive lemma-selection algorithm (used also by epcllemma):
\begin{algorithm}[!ht]
\caption{Best lemmas}
\begin{algorithmic}[1]
\Require{a lemma-quality metric $Q$, set of lemmas $Lemmas$, an initial set of named lemmas $Named_0 \subset Lemmas$, and a required number of lemmas $M$}
\Ensure{set $Named$ of $M$ best lemmas according to $Q$}
\State $Named \gets Named_0$
\State $m \gets 0$
  \While{$m < M$}
    \For{$i \in Lemmas$}
      \State \Call{Calculate}{$Q_{Named}(i)$}
    \EndFor
    \State $j \gets argmax \{ Q_{Named}(i) : i \in Lemmas \setminus Named \}$
    \State $Named \gets Named \cup \{j\}$
    \State $m \gets m+1$
  \EndWhile
  \State \Call{Return}{$Named$}
\end{algorithmic}
\end{algorithm}

There are two possible choices of $Named_0$: either the empty set, or
the set of all human-named theorems. This choice depends on whether we
want re-organize the library from scratch, or whether we just want to
select good lemmas that complement the human-named theorems. Below we
experiment with both approaches. Note that this algorithm is currently
quite expensive: the fast epcllemma implementation takes 65 seconds to
update the lemma qualities over the whole Flyspeck graph after each
change of the $Named$ set. This means that producing the first 10000
Flyspeck lemmas takes 180 CPU hours. That is why most of the
experiments are limited to the core HOL Light graph where this
takes about 1 second and 3 hours respectively.

\section{Evaluation Scenarios and Issues}
\label{Scenarios}
To assess and develop the lemma-mining methods we define several
evaluation scenarios that vary in speed, informativeness and
rigor. The simplest and least rigorous is the \textit{expert-evaluation} scenario:
We use our knowledge of the formal corpora to quickly see if the
top-ranked lemmas produced by a particular method look
plausible. Because of its size, this is the only evaluation done for
the whole Flyspeck corpus so far.

The \textit{cheating ATP} scenario uses the full proof graph of a
corpus to compute the set of the (typically 10,000) best lemmas
($BestLemmas$) for the whole corpus. Then the set of newly named
theorems ($NewThms$) is defined as the union of $BestLemmas$ with the
set of originally named theorems ($OrigThms$): $NewThms \coloneqq
BestLemmas \cup OrigThms$. The derived graph $G_{NewThms}$ of direct
dependencies among the elements of $NewThms$ is used for ATP
evaluation, which may be done in two ways: with human selection and
with AI selection. When using human selection, we try to prove each
lemma from its parents in $G_{NewThms}$. When using AI selection, we
use the chronological order (see Section~\ref{Approach}) of $NewThms$
to incrementally train and evaluate the $k$-NN machine learner~\cite{EasyChair:74} on the
direct dependencies from $G_{NewThms}$. This produces for each new
theorem an ATP problem with premises advised by the learner trained on
the $G_{NewThms}$ dependencies of the preceding new theorems. This
scenario may do a lot of cheating, because when measuring the ATP
success on $OrigThms$, a particular theorem $i$ might be proved with
the use lemmas from $NewThms$ that have been stated for the first time
only in the original proof of $i$ (we call such lemmas
\textit{directly preceding}). In other words, such lemmas did not
exist before the original proof of $i$ was started, so they could not
possibly be suggested by lemma-quality metrics for proving $i$. Such
directly preceding lemmas could also be very close to $i$, and thus
equally hard to prove.

The \textit{almost-honest ATP} scenario is like the \textit{cheating
  ATP} scenario, however directly preceding new lemmas are replaced by
their closest $OrigThms$ ancestors.  This scenario is still not fully honest,
because the lemmas are computed according to their lemma quality
measured on the full proof graph. In particular, when proving an early
theorem $i$ from $OrigThms$, the newly used parents of $i$ are lemmas
whose quality was clear only after taking into account the theorems
that were proved later than $i$. These theorems and their proofs
however did not exist at the time of proving $i$. Still, we consider
this scenario sufficiently honest for most of the ATP evaluations done with
over the whole core HOL Light dataset. 

The \textit{fully-honest ATP} scenario removes this last objection, at
the price of using considerably more resources for a single evaluation. For
each originally named theorem $j$ we limit the proof graph used for
computing $BestLemmas$ to the proofs that preceded $j$. Since
computing $BestLemmas$ for the whole core HOL Light takes at least
three hours for the $Q_i$ and $EQ_i$ methods, the full evaluation on
all 1,984 core HOL Light theorems would take about 2,000 CPU
hours. That is why we further scale down this evaluation by doing it
only for every tenth theorem in core HOL Light.

The \textit{chained-conjecturing ATP} scenario is similar to the
cheating scenario, but with limits imposed on the directly preceding
lemmas. In $chain_1$\textit{-conjecturing}, any (possibly directly
preceding) lemma used to prove a theorem $i$ must itself have an ATP
proof using only $OrigThms$. In other words, it is allowed to guess
good lemmas that still do not exist, but such lemmas must not be hard
to prove from $OrigThms$. Analogously for
$chain_2$\textit{-conjecturing} (resp. $chain_N$), where lemmas
provable from $chain_1$-lemmas (resp. $chain_{N-1}$) are allowed to be
guessed.  To some extent, this scenario measures the theoretical ATP
improvement obtainable with guessing of good intermediate lemmas.

\section{Experiments}
\label{Experiments}
The ATP experiments are done on the same hardware and using the same
setup that was used for the earlier evaluations described
in~\cite{holyhammer,EasyChair:74}: All systems are run with 30s time
limit on a 48-core server with AMD Opteron 6174 2.2 GHz CPUs, 320 GB
RAM, and 0.5 MB L2 cache per CPU.  When using only the original
theorems, the success rate of the 14 most complementary AI/ATP methods
run with 30s time limit each and restricted to the 1954 core HOL Light
theorems is 65.2\% (1275 theorems) and the union of all methods solves
65.4\% (1278 theorems). 

In the very optimistic \textit{cheating} scenario (limited only to the
$Q_i$ metrics), this numbers go up to 76.5\% (1496 theorems)
resp. 77.9\% (1523 theorems). As mentioned in Section~\ref{Scenarios},
many proofs in this scenario may however be too simple because a close
directly preceding lemma was used by the
lemma-mining/machine-learning/ATP stack. This became easy to see already
when using the \textit{almost-honest} scenario, where the 14 best
methods (including also $EQ_i$ and $PR_i$) solve together only 66.3\%
(1296 theorems) and the union of all methods solves 68.9\% (1347
theorems). The resource-intensive \textit{fully-honest} evaluation is
limited to a relatively small subset of the core HOL Light theorems,
however it confirms the \textit{almost-honest} results. While the
original success rate was 61.7\% (less than 14 methods are needed to
reach it), the success rate with lemma mining went up to 64.8\%
(again, less than 14 methods are needed). This means that the
non-cheating lemma-mining approaches so far improve the overall
performance of the AI/ATP methods over core HOL Light by about 5\%.
The best method in the \textit{fully-honest} evaluation is $Q_2$ which
solves 46.2\% of the original problems when using 512 premises,
followed by $EQ_2$ (using the longest inference chain instead of $D$),
which solves 44.6 problems also with 512 premises. The best
PageRank-based method is $PR_2$ (PageRank divided by size), solving
41.4\% problems with 128 premises.

An interesting middle-way between the cheating and non-cheating
scenarios is the \textit{chained-conjecturing} evaluation, which
indicates the possible improvement when guessing good lemmas that are
``in the middle'' of long proofs. Since this is also quite expensive,
only the best lemma-mining method ($Q_2$) was evaluated so far. $Q_2$
itself solves (altogether, using different numbers of premises) 54.5\%
(1066) of the problems. This goes up to 61.4\% (1200 theorems) when
using only $chain_1$\textit{-conjecturing} and to 63.8\% (1247
theorems) when allowing also $chain_2$ and
$chain_3$\textit{-conjecturing}. These are 12.6\% and
17.0\% improvements respectively.

Finally, since regular lemma-mining/machine-learning/ATP evaluations over the whole Flyspeck corpus are still outside our resources, we present below several best lemmas computed by epcllemma's
$EQ_2$ method over the 97,714,465-node-large proof graph of all Flyspeck lemmas:\footnote{\url{http://mizar.cs.ualberta.ca/~mptp/lemma_mining/proofs.grf1.lm.flfull}} 

\begin{verbatim}
|- a + c + d = c + a + d
|- x * (y + z) = x * y + x * z
|- (a + b) + c = a + b + c
|- &1 > &0
|- a ==> b <=> ~a \/ b
|- BIT1 m + BIT0 n = BIT1 (m + n)
\end{verbatim}

\section{Future Work and Conclusion}
\label{Future}

We have proposed, implemented and evaluated several approaches that try to efficiently
find the best lemmas and re-organize a large corpus of computer-understandable human
mathematical ideas, using the millions of logical dependencies between
the corpus' atomic elements. We believe that such conceptual
re-organization is a very interesting AI topic that is best studied in the context of large, fully
semantic corpora such as HOL Light and Flyspeck. 
The byproduct of this work
are the exporting and post-processing techniques resulting in the
publicly available proof graphs that can serve as a basis for further
research.

The most conservative improvement in the strength of automated reasoning
obtained so far over the core HOL Light thanks to lemma mining is about
5\%. There are potential large improvements if the guessing of lemmas is
improved. The benefits from lemma-mining should be larger when proving over
larger corpora and when proving larger steps, but a number of implementational
issues need to be addressed to scale the lemma-mining methods to very large
corpora such as Flyspeck.

There are many further directions for this work. The lemma-mining methods
can be made faster and more incremental, so that the lemma quality is not
completely recomputed after a lemma is named. Fast PageRank-based clustering
should be efficiently implemented and possibly combined with the other methods
used. ATP-style normalizations such as subsumption need to be correctly merged
with the detailed level of inferences used by the HOL Light proof graph. The
whole approach could also be implemented on a higher level of inferences, using
for example the granularity corresponding to time-limited MESON ATP steps.
Guessing of good intermediate lemmas for proving harder theorems is an obvious
next step, the value of which has already been established to a certain extent in
this work.

\section{Acknowledgments}

We would like to thank Stephan Schulz for help with running epcllemma,
Yury Puzis and Geoff Sutcliffe for their help with the Agint tool and
Ji\v{r}\'i Vysko\v{c}il and Petr Pudl\'ak for many discussions about
extracting interesting lemmas from proofs.

\bibliography{ate11}

\begin{thebibliography}{10}

\bibitem{abs-1108-3446}
Jesse Alama, Tom Heskes, Daniel K\"{u}hlwein, Evgeni Tsivtsivadze, and Josef
  Urban.
\newblock Premise selection for mathematics by corpus analysis and kernel
  methods.
\newblock {\em Journal of Automated Reasoning}, 2013.
\newblock \url{http://dx.doi.org/10.1007/s10817-013-9286-5}.

\bibitem{AlamaKU12}
Jesse Alama, Daniel K{\"u}hlwein, and Josef Urban.
\newblock {Automated and Human Proofs in General Mathematics: An Initial
  Comparison}.
\newblock In Nikolaj Bj{\o}rner and Andrei Voronkov, editors, {\em LPAR},
  volume 7180 of {\em LNCS}, pages 37--45. Springer, 2012.

\bibitem{AvrachenkovDNPS08}
Konstantin Avrachenkov, Vladimir Dobrynin, Danil Nemirovsky, Son~Kim Pham, and
  Elena Smirnova.
\newblock Pagerank based clustering of hypertext document collections.
\newblock In Sung-Hyon Myaeng, Douglas~W. Oard, Fabrizio Sebastiani, Tat-Seng
  Chua, and Mun-Kew Leong, editors, {\em SIGIR}, pages 873--874. ACM, 2008.

\bibitem{DBLP:conf/itp/2013}
Sandrine Blazy, Christine Paulin-Mohring, and David Pichardie, editors.
\newblock {\em Interactive Theorem Proving - 4th International Conference, ITP
  2013, Rennes, France, July 22-26, 2013. Proceedings}, volume 7998 of {\em
  Lecture Notes in Computer Science}. Springer, 2013.

\bibitem{mizar-in-a-nutshell}
Adam Grabowski, Artur Korni{\l}owicz, and Adam Naumowicz.
\newblock {M}izar in a nutshell.
\newblock {\em Journal of Formalized Reasoning}, 3(2):153--245, 2010.

\bibitem{Hales05}
Thomas~C. Hales.
\newblock Introduction to the {F}lyspeck project.
\newblock In Thierry Coquand, Henri Lombardi, and Marie-Fran\c{c}oise Roy,
  editors, {\em Mathematics, Algorithms, Proofs}, volume 05021 of {\em Dagstuhl
  Seminar Proceedings}. Internationales Begegnungs- und Forschungszentrum
  f{\"u}r Informatik (IBFI), Schloss Dagstuhl, Germany, 2005.

\bibitem{Harrison96}
John Harrison.
\newblock {HOL Light}: A tutorial introduction.
\newblock In Mandayam~K. Srivas and Albert~John Camilleri, editors, {\em
  FMCAD}, volume 1166 of {\em LNCS}, pages 265--269. Springer, 1996.

\bibitem{HoderV11}
Krystof Hoder and Andrei Voronkov.
\newblock Sine qua non for large theory reasoning.
\newblock In Nikolaj Bj{\o}rner and Viorica Sofronie-Stokkermans, editors, {\em
  CADE}, volume 6803 of {\em LNCS}, pages 299--314. Springer, 2011.

\bibitem{KaliszykK13}
Cezary Kaliszyk and Alexander Krauss.
\newblock Scalable {LCF}-style proof translation.
\newblock In Blazy et~al. \cite{DBLP:conf/itp/2013}, pages 51--66.

\bibitem{holyhammer}
Cezary Kaliszyk and Josef Urban.
\newblock Learning-assisted automated reasoning with {F}lyspeck.
\newblock {\em CoRR}, abs/1211.7012, 2012.

\bibitem{KaliszykU13}
Cezary Kaliszyk and Josef Urban.
\newblock Automated reasoning service for {HOL Light}.
\newblock In Jacques Carette, David Aspinall, Christoph Lange, Petr Sojka, and
  Wolfgang Windsteiger, editors, {\em MKM/Calculemus/DML}, volume 7961 of {\em
  Lecture Notes in Computer Science}, pages 120--135. Springer, 2013.

\bibitem{EasyChair:74}
Cezary Kaliszyk and Josef Urban.
\newblock Stronger automation for {F}lyspeck by feature weighting and strategy
  evolution.
\newblock In Jasmin~Christian Blanchette and Josef Urban, editors, {\em PxTP
  2013}, volume~14 of {\em EPiC Series}, pages 87--95. EasyChair, 2013.

\bibitem{KuhlweinBKU13}
Daniel K{\"u}hlwein, Jasmin~Christian Blanchette, Cezary Kaliszyk, and Josef
  Urban.
\newblock {MaSh}: Machine learning for {S}ledgehammer.
\newblock In Blazy et~al. \cite{DBLP:conf/itp/2013}, pages 35--50.

\bibitem{KuhlweinSU13}
Daniel K{\"u}hlwein, Stephan Schulz, and Josef Urban.
\newblock {E-MaLeS} 1.1.
\newblock In Maria~Paola Bonacina, editor, {\em CADE}, volume 7898 of {\em
  Lecture Notes in Computer Science}, pages 407--413. Springer, 2013.

\bibitem{KuhlweinLTUH12}
Daniel K{\"u}hlwein, Twan van Laarhoven, Evgeni Tsivtsivadze, Josef Urban, and
  Tom Heskes.
\newblock Overview and evaluation of premise selection techniques for large
  theory mathematics.
\newblock In Bernhard Gramlich, Dale Miller, and Uli Sattler, editors, {\em
  IJCAR}, volume 7364 of {\em LNCS}, pages 378--392. Springer, 2012.

\bibitem{McC-Prover9-URL}
William McCune.
\newblock {Prover9 and Mace4}.
\newblock \url{http://www.cs.unm.edu/~mccune/prover9/}, 2005--2010.

\bibitem{MengP08}
Jia Meng and Lawrence~C. Paulson.
\newblock Translating higher-order clauses to first-order clauses.
\newblock {\em J. Autom. Reasoning}, 40(1):35--60, 2008.

\bibitem{page98pagerankTechreport}
Lawrence Page, Sergey Brin, Rajeev Motwani, and Terry Winograd.
\newblock The {PageRank} citation ranking: Bringing order to the {Web}.
\newblock Technical report, Stanford Digital Library Technologies Project,
  1998.

\bibitem{Pud06-ESCoR}
Petr Pudl\'{a}k.
\newblock Search for faster and shorter proofs using machine generated lemmas.
\newblock In G.~Sutcliffe, R.~Schmidt, and S.~Schulz, editors, {\em
  {Proceedings of the FLoC'06 Workshop on Empirically Successful Computerized
  Reasoning, 3rd International Joint Conference on Automated Reasoning}},
  volume 192 of {\em CEUR Workshop Proceedings}, pages 34--52, 2006.

\bibitem{PuzisGS06}
Yury Puzis, Yi~Gao, and Geoff Sutcliffe.
\newblock Automated generation of interesting theorems.
\newblock In Geoff Sutcliffe and Randy Goebel, editors, {\em FLAIRS
  Conference}, pages 49--54. AAAI Press, 2006.

\bibitem{Vampire}
Alexandre Riazanov and Andrei Voronkov.
\newblock The design and implementation of {VAMPIRE}.
\newblock {\em AI Commun.}, 15(2-3):91--110, 2002.

\bibitem{Sch00}
Stephan Schulz.
\newblock {\em {Learning search control knowledge for equational deduction}},
  volume 230 of {\em DISKI}.
\newblock Infix Akademische Verlagsgesellschaft, 2000.

\bibitem{Sch02-AICOMM}
Stephan Schulz.
\newblock {E - A Brainiac Theorem Prover}.
\newblock {\em AI Commun.}, 15(2-3):111--126, 2002.

\bibitem{Sut01-LPAR}
Geoff Sutcliffe.
\newblock {The Design and Implementation of a Compositional
  Competition-Cooperation Parallel ATP System}.
\newblock In H.~de~Nivelle and S.~Schulz, editors, {\em {Proceedings of the 2nd
  International Workshop on the Implementation of Logics}}, number
  MPI-I-2001-2-006 in Max-Planck-Institut f{\"u}r Informatik, Research Report,
  pages 92--102, 2001.

\bibitem{SutcliffeP07}
Geoff Sutcliffe and Yury Puzis.
\newblock {SRASS} - a semantic relevance axiom selection system.
\newblock In Frank Pfenning, editor, {\em CADE}, volume 4603 of {\em LNCS},
  pages 295--310. Springer, 2007.

\bibitem{Urb04-MPTP0}
Josef Urban.
\newblock {MPTP - Motivation, Implementation, First Experiments}.
\newblock {\em Journal of Automated Reasoning}, 33(3-4):319--339, 2004.

\bibitem{Urban06-ijait}
Josef Urban.
\newblock {M}o{M}{M} - fast interreduction and retrieval in large libraries of
  formalized mathematics.
\newblock {\em Int. J. on Artificial Intelligence Tools}, 15(1):109--130, 2006.

\bibitem{blistr}
Josef Urban.
\newblock {BliStr: The Blind Strategymaker}.
\newblock {\em CoRR}, abs/1301.2683, 2013.

\bibitem{US+08}
Josef Urban, Geoff Sutcliffe, Petr Pudl{\'a}k, and Ji\v{r}\'{\i} Vysko\v{c}il.
\newblock {MaLARea SG1 - Machine Learner for Automated Reasoning with Semantic
  Guidance}.
\newblock In Alessandro Armando, Peter Baumgartner, and Gilles Dowek, editors,
  {\em IJCAR}, volume 5195 of {\em LNCS}, pages 441--456. Springer, 2008.

\bibitem{UrbanV13}
Josef Urban and Ji\v{r}\'{\i} Vysko\v{c}il.
\newblock Theorem proving in large formal mathematics as an emerging {AI}
  field.
\newblock In Maria~Paola Bonacina and Mark~E. Stickel, editors, {\em Automated
  Reasoning and Mathematics: Essays in Memory of William McCune}, volume 7788
  of {\em LNAI}, pages 240--257. Springer, 2013.

\bibitem{UrbanVS11}
Josef Urban, Ji\v{r}\'{\i} Vysko\v{c}il, and Petr \v{S}t\v{e}p{\'a}nek.
\newblock {MaLeCoP}: Machine learning connection prover.
\newblock In Kai Br{\"u}nnler and George Metcalfe, editors, {\em TABLEAUX},
  volume 6793 of {\em LNCS}, pages 263--277. Springer, 2011.

\bibitem{Veroff96}
Robert Veroff.
\newblock Using hints to increase the effectiveness of an automated reasoning
  program: Case studies.
\newblock {\em J. Autom. Reasoning}, 16(3):223--239, 1996.

\bibitem{WenzelPN08}
Makarius Wenzel, Lawrence~C. Paulson, and Tobias Nipkow.
\newblock The {I}sabelle framework.
\newblock In Otmane~A\"{\i}t Mohamed, C{\'e}sar~A. Mu{\~n}oz, and Sofi{\`e}ne
  Tahar, editors, {\em TPHOLs}, volume 5170 of {\em Lecture Notes in Computer
  Science}, pages 33--38. Springer, 2008.

\bibitem{WO+84}
Larry Wos, Ross Overbeek, Ewing~L. Lusk, and Jim Boyle.
\newblock {\em {Automated Reasoning: Introduction and Applications}}.
\newblock Prentice-Hall, 1984.

\end{thebibliography}
\bibliographystyle{plain}

\end{document}